\documentclass[letterpaper, 10 pt, conference]{ieeeconf}  

\IEEEoverridecommandlockouts
\usepackage[vlined, ruled, boxed, linesnumbered]{algorithm2e}

\SetCommentSty{mycommfont}

\SetKwProg{Function}{}{}{}

\usepackage{graphicx,subfig}
\usepackage{graphics}
\usepackage{amssymb,amsmath}
\usepackage{amssymb}
\usepackage[noend]{algpseudocode}
\usepackage[font={small}]{caption}

\usepackage[usenames, dvipsnames]{color}

\usepackage[normalem]{ulem}

\usepackage{tikz}
\usepackage{textcomp}
\usepackage{lipsum}

\usepackage{booktabs}

\newcommand{\cbf}{\mathbf{c}}

\newcommand{\pbf}{\mathbf{p}}

\newcommand{\zbf}{\mathbf{z}}

\newcommand{\Tbf}{\mathbf{T}}

\newcommand{\Ccal}{\mathcal{C}}

\newcommand{\Ecal}{\mathcal{E}}

\newcommand{\Gcal}{\mathcal{G}}

\newcommand{\Ical}{\mathcal{I}}

\newcommand{\Lcal}{\mathcal{L}}

\newcommand{\Pcal}{\mathcal{P}}

\newcommand{\Rcal}{\mathcal{R}}
\newcommand{\Scal}{\mathcal{S}}

\newcommand{\Vcal}{\mathcal{V}}

\newcommand{\Zcal}{\mathcal{Z}}

\DeclareMathOperator*{\argmin}{arg\,min}

\title{\LARGE \bf
Fusing Concurrent Orthogonal Wide-aperture Sonar Images for \\
Dense Underwater 3D Reconstruction
}

\author{John McConnell, John D. Martin and Brendan Englot
\thanks{J. McConnell, J.D. Martin and B. Englot are with the Department of Mechanical Engineering, Stevens Institute of Technology,
        Hoboken, NJ 07030, USA
        {\tt\small$\{$jmcconn1,jmarti3,benglot$\}$@stevens.edu}}%
}

\begin{document}

\maketitle
\thispagestyle{empty}
\pagestyle{empty}

\begin{abstract}
We propose a novel approach to handling the ambiguity in elevation angle associated with the observations of a forward looking multi-beam imaging sonar, and the challenges it poses for performing an accurate 3D reconstruction. We utilize a pair of sonars with orthogonal axes of uncertainty to independently observe the same points in the environment from two different perspectives, and associate these observations. Using these concurrent observations, we can create a dense, fully defined point cloud at every time-step to aid in reconstructing the 3D geometry of underwater scenes. We will evaluate our method in the context of the current state of the art, for which strong assumptions on object geometry limit applicability to generalized 3D scenes. We will discuss results from laboratory tests that quantitatively benchmark our algorithm's reconstruction capabilities, and results from a real-world, tidal river basin which qualitatively demonstrate our ability to reconstruct a cluttered field of underwater objects.
\end{abstract}

\section{Introduction}
Over the past several years, autonomous underwater vehicles (AUVs) have seen increased usage, addressing needs ranging from ship inspection to the assessment of offshore oil and gas assets in deep water \cite{SLB-2020}. To perform these tasks, AUVs can be equipped with a variety of sensors to localize and perform asset inspection. However, when operating in turbid, dark conditions, cameras lose their viability, making sonars the perceptual sensor of choice. When it comes to sonar, three primary modalities have proven useful in a wide variety of applications. Firstly, side-scan sonar has achieved great utility and ubiquity in seafloor mapping applications. Secondly, profiling and bathymetry sonars provide a narrow beam that is highly accurate, but requires many samples to achieve coverage. 
Thirdly, wide-aperture, forward-looking multi-beam imaging sonars provide an expansive field of view that may be flexibly tasked to gather imagery from a variety of perspectives at a fraction of the cost of its narrow beam competitors. 
However, imaging sonar is characterized by a high signal to noise ratio and under-constrained measurements, providing flattened 2D imagery of an observed 3D volume. Thus, a subsequent challenge is performing accurate 3D reconstruction of observed objects using its measurements, which lack an elevation angle.

\begin{figure}[t]
\centering
\subfloat[BlueROV2 system overview\label{fig:1a}]{\includegraphics[height=4.8cm]{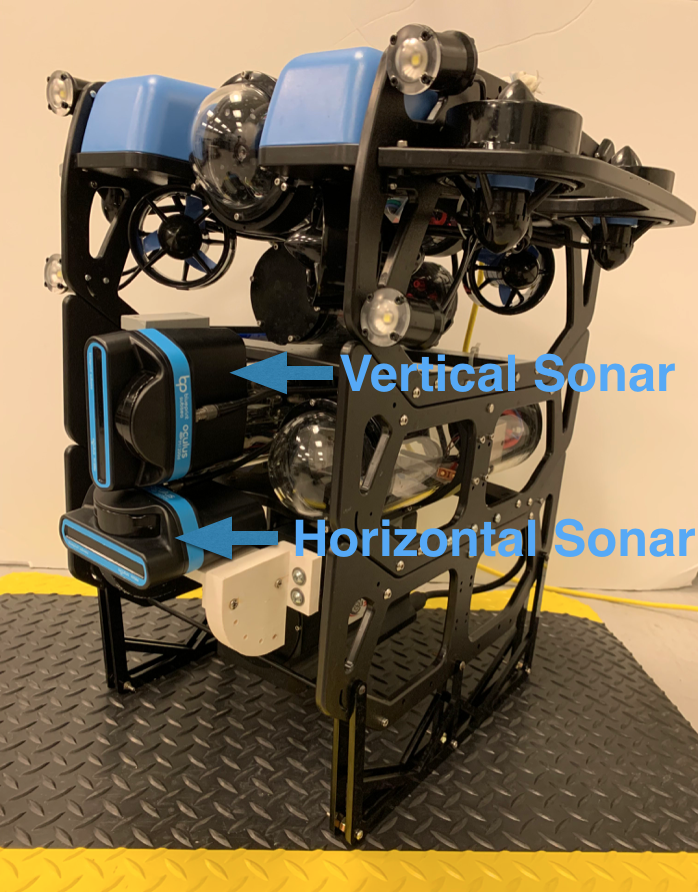}}\
\subfloat[Sonar overlapping fields of view]{\includegraphics[height=4.2cm]{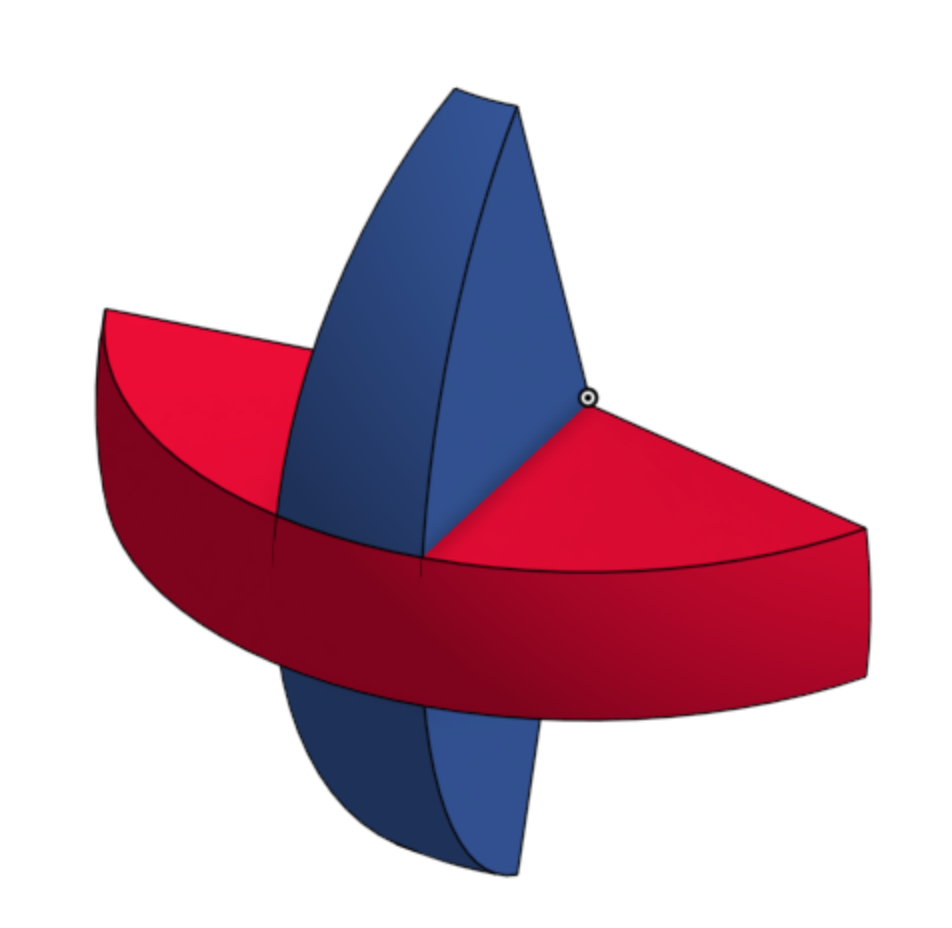}}\\
\subfloat[Reconstructed Pier Pilings ]{\includegraphics[height=3.2cm,width=0.49\linewidth]{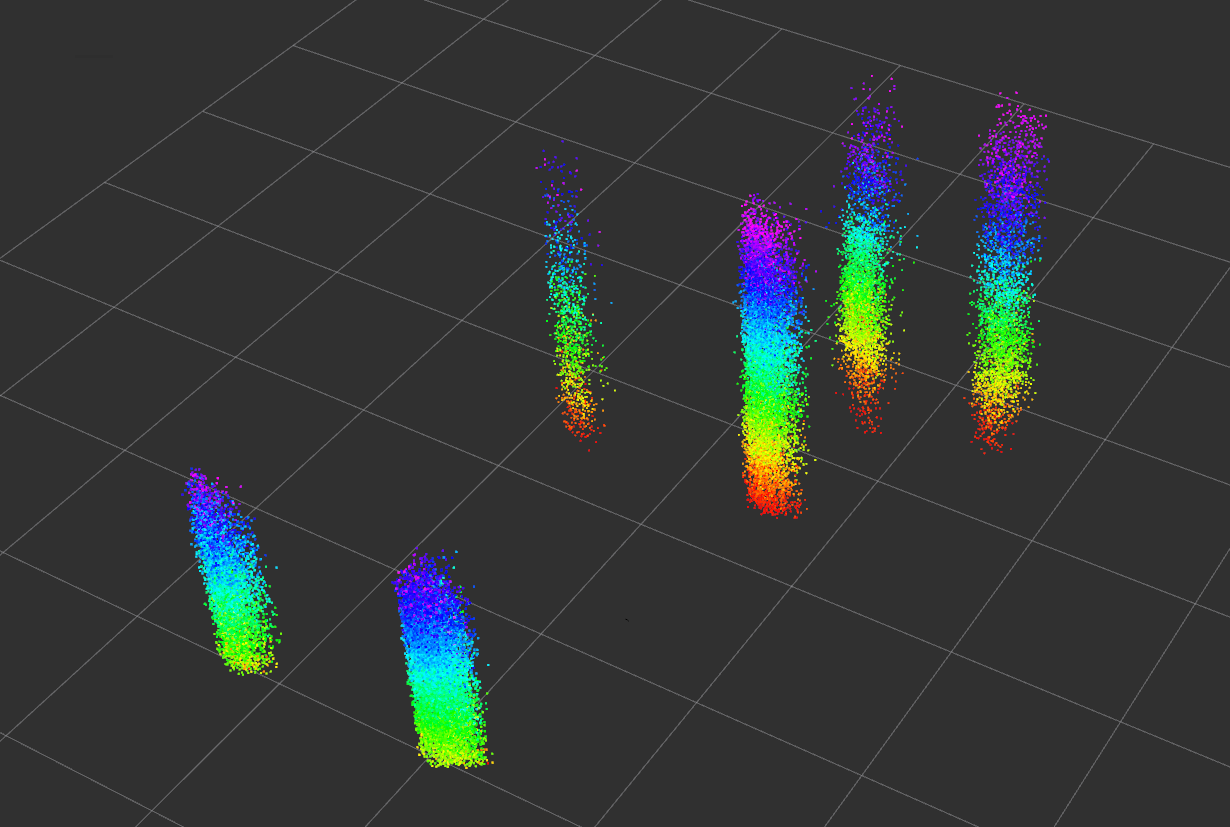}}\

\caption{\textbf{System Overview.} Using two multi-beam sonars, correspondences between their observations are computed to extract 3D point clouds. Sonar fields of view corresponding to the hardware arrangement in (a) are shown in (b) - the red swath is from the horizontal sonar and the blue is from the vertical sonar, shown at a range of 10m. Fig 1(c) shows a reconstruction of pier pilings in the Hudson River. Data was collected at a fixed depth.}
\vspace{-5mm}
\label{fig:1}
\end{figure}

In this work, we will focus on the capabilities of AUVs operating in cluttered environments, such as subsea oil fields and nearshore piers where profiling sonars cannot provide the requisite situational awareness. While profiling sonars, paired with suitable AUV state estimation, can provide highly accurate 3D reconstructions, a profiling sonar can only observe a narrow slice of the environment at a time, making the problem of navigating in three-dimensional clutter potentially intractable.  In this work we are motivated by the problem of providing an AUV with the best-possible situational awareness to safely navigate in clutter, rather than computing the most accurate reconstruction possible. 

We propose to address an imaging sonar's lack of elevation angle by using an array of two orthogonally oriented sonars (Fig. \ref{fig:1}). Our goal, for AUVs to operate reliably in cluttered 3D environments, requires dense volumetric data to be extracted from the perceptual system at every time step, supporting localization and collision avoidance. Accordingly, we propose a novel method to compute the correspondences between two overlapping sonar images of the same scene collected from different vantage points. The core contribution of this paper is a methodology for identifying suitable features and their correspondences across pairs of orthogonal sonar images, and hence measuring the features' locations in 3D Euclidean space. The outcome is the ability to perform dense 3D reconstruction using wide-aperture multi-beam imaging sonar, while making no restrictive assumptions about the scenes in view. Moreover, we do not rely on re-observing features at a later time step to resolve the ambiguity, representing progress towards being able to reconstruct complex 3D cluttered, dynamic scenes with wide-aperture imaging sonar.

In the sections to follow, we will first discuss related work that we draw on for inspiration and benchmarking. Next, we will discuss the specific challenges associated with 3D reconstruction with sonar and precisely define the problem to be solved. Lastly, we will present three experiments. The first will show that our algorithm is comparable with the state of the art in simple cases. The second is a challenging case that violates the assumptions of previous work in this area. Lastly, a field demonstration that shows our algorithm works at scale when observing a cluttered field of objects.  

\begin{figure}[]
\centering
\includegraphics[width=\columnwidth]{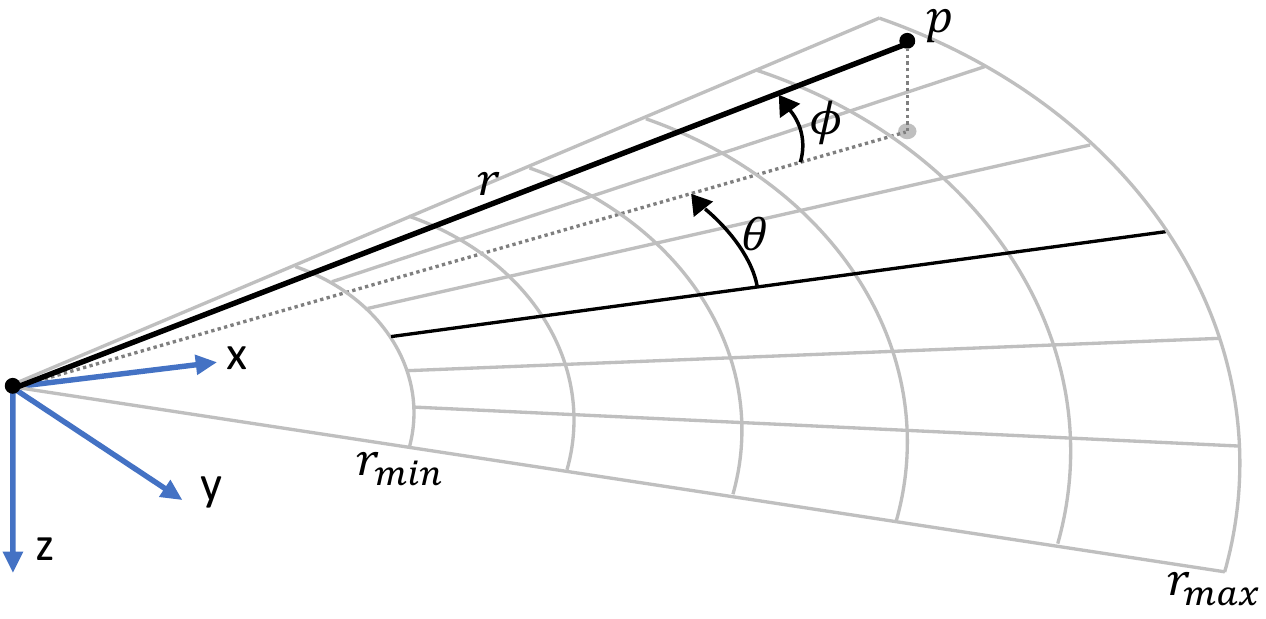}
\caption{\textbf{Forward looking imaging sonar model.} The point $p$ can be represented by $[r,\theta,\phi]^T$ in a spherical coordinate frame. The range $r$ and the bearing angle $\theta$ of $p$ are measured, while the elevation angle $\phi$ is not captured in the resulting 2D sonar image.}
\vspace{-3mm}
\label{fig:son_geo}
\end{figure}

\section{Related Work}
\subsection{Estimating Elevation Angle}

The challenges associated with wide-aperture multi-beam imaging sonars have inspired an impressive body of work to address the fundamental limitations of their under-constrained measurements. Firstly, work from Aykin \cite{Aykin-2013, Aykin-2013-1} estimates the elevation angle of sonar image pixels in scenes where objects are lying on the seafloor.

The recent work of Westman \cite{Westman-2019} extends the work of Aykin and shows excellent results in a constrained nearshore pier environment. 
However, these methods rely on several assumptions that may often be violated. Firstly, both \cite{Aykin-2013-1} and \cite{Westman-2019} assume that all objects in view have their range returns monotonically increase or decrease with elevation angle. While this assumption may hold true for some objects, it hinders the application of their methods to arbitrary objects and scenes. Additionally, \cite{Aykin-2013-1} requires the leading and trailing edge of an observed object; this is obtained by examining the shadow area behind a segmentation created by the sonar's downward grazing angle. In contrast, \cite{Westman-2019} only needs to identify the leading \textit{or} trailing edge of the object; however, in the experiments
shown, the leading edge is always the closest return because of the sonar's downward grazing angle. Using a downward grazing angle makes the problem significantly simpler to solve, but comes at a price. By tilting the sonar downward, making the upper edge of the sonar beam parallel with the water plane, the AUV's situational awareness may be hampered. In cluttered environments, an AUV would be unable to see above it before transiting upward, and a safe navigation solution may not always be possible.  Moreover, if the vehicle is perturbed in a way that violates this geometric assumption, the perception system could be driven to inaccuracy. 

We also note that a recent technique has employed deep learning with convolutional neural networks to estimate the elevation angle associated with imaging sonar observations \cite{DeBortoli-2019}. In our paper, we assume a vehicle may lack opportunities for prior training and exposure to the subsea objects and scenes it may encounter in a given mission.  

\subsection{Carving Out Low-Intensity Background Pixels}

Another method proposed by Aykin \cite{Aykin-2017} applies a space carving approach to produce surface models from an image's low-intensity background, which outer-bound the objects of interest. The min-filtering voxel grid modeling approach from Guerneve \cite{Guerneve-2018} similarly removes voxels from an object model based on observations of low-intensity pixels. These approaches require the objects of interest to be observed from multiple vantage points to achieve accurate reconstructions.

\begin{figure}[t]
\centering
\includegraphics[width=\columnwidth]{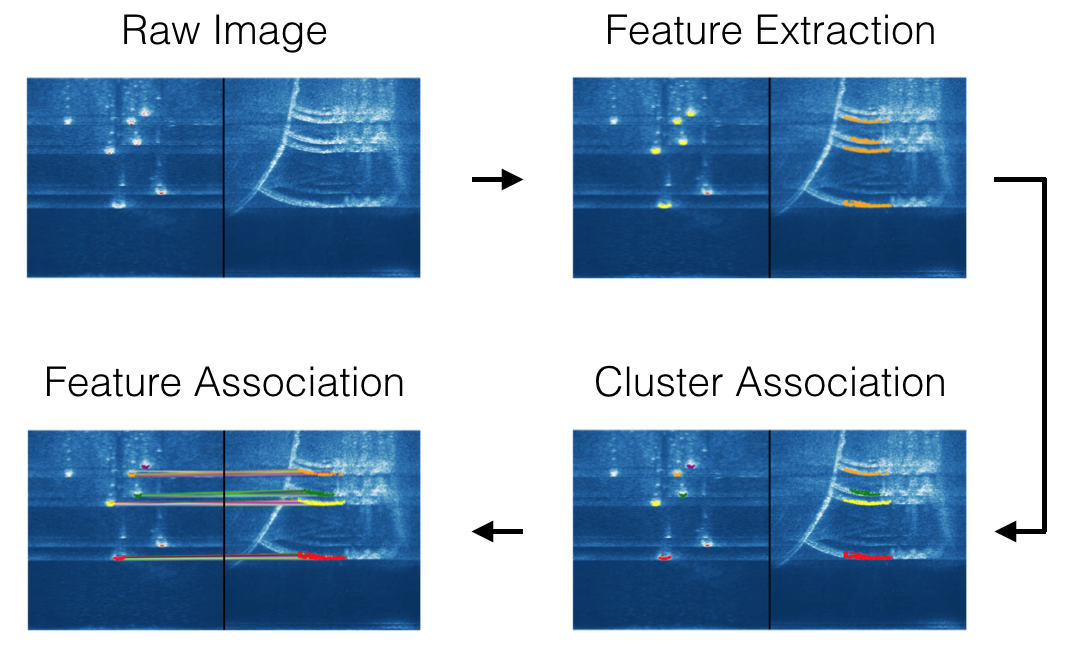}
\caption{\textbf{System architecture overview.} Raw image pairs are taken from the same time step, features are extracted, the features are clustered and then matched using cluster labels as constraints.}
\vspace{-3mm}
\label{fig:fig3}
\end{figure}

\subsection{Acoustic Structure From Motion}

A similar core issue can also be addressed from a simultaneous localization and mapping (SLAM) perspective. Rather than trying to estimate the elevation of pixels in a single frame, these works acquire features and use a series of views combined with a pose graph back-end \cite{Kaess-2008} to determine 3D structure. This was proposed by Huang \cite{Huang-2016} in acoustic structure from motion (ASFM). This initial implementation has limitations, chief of which is the reliance on manually extracted features. This work was later built on by Wang \cite{Wang-2019}, incorporating automated feature extraction and tracking. While these methods provide impressive results, they are focused on reconstructing the terrain under the vehicle, rather than providing adequate situational awareness around the vehicle.  The limitation of these methods for AUVs in clutter is the perception system requiring a series of frames to recover 3D information, rather than a single timestep. 

\subsection{Stereo Imagery and Feature Correspondences}

Lastly, much has been accomplished in the realm of stereo vision, and of specific relevance, computing the correspondences between two vantage points of the same scene. 
This concept is widely examined in an extensive body of literature [11-14].
Further, many feature extraction methods have been developed over the years, including SIFT \cite{Lowe-2004}, SURF \cite{Bay-2007}, ORB \cite{Rublee-2011} and KAZE \cite{Alcantarilla-2012}. Recently AKAZE \cite{Alcantarilla-2013} has shown promise in computing correspondences between acoustic images from a multi-beam imaging sonar. Westman \cite{Westman-2018} shows the utility of AKAZE features in a SLAM solution using acoustic imagery. Further, Wang \cite{Wang-2019} utilizes these same features in terrain reconstruction with acoustic imagery. 

We also note that the specific concept of using two imaging sonars in stereo for 3D perception has been employed previously, but with relatively small differences in position and orientation between the sonars, and for reconstructing \textit{sparse} sets of point features. The concept, first proposed by Assalih \cite{Assalih-2009}, was implemented and further analyzed by Negahdaripour \cite{Negahdaripour-2018}, and used to build 3D maps of both sparse features extracted from a planar grid, and from small seafloor objects. Beyond its output of sparse features, this work stands in contrast to ours as the sensors used have aligned axes of uncertainty, rather than the orthogonal axes of uncertainty utilized in our work. 
A notable example of \textit{dense} 3D mapping is the Sparus AUV's two orthogonally oriented single-beam mechanically scanning sonars (one imager and one profiler), used for cave mapping \cite{Mallios-2016}, \cite{Mallios-2017}. However, the imager and profiler were used independently to address SLAM and 3D mapping, respectively, in separate steps.

\section{Problem Description}
We consider the problem of reconstructing 3D geometry with data gathered by an imaging sonar. Environments are represented as a collection of points $\pbf \in \mathbb{R}^3$, which define the location of its surface relative to a robot. We express these using coordinates from the robot's local frame $\Rcal$:
\begin{align}
    \pbf^{(\Rcal)} &= \begin{pmatrix}  X \\  Y \\  Z \end{pmatrix} 
    = R\begin{pmatrix} \cos{\phi} \cos{\theta} \\ 
    \cos{\phi}\sin{\theta} \\ 
    \sin{\phi} \end{pmatrix}.
    \label{eq:tx_to_cartesian}
\end{align}
Here, $X,Y,$ and $Z$ are the Cartesian coordinates corresponding to the range $R\in \mathbb{R}_+$, bearing $\theta\in \Theta$, and elevation $\phi \in \Phi$, with $\Theta,\Phi \subseteq [-\pi,\pi)$, illustrated in Fig. \ref{fig:son_geo}. An imaging sonar measures points in spherical coordinates by emitting acoustic pulses and measuring the associated intensity $\gamma \in \mathbb{R}_+$ from their returns. This information is organized into an \textit{intensity image}, 
which we view as a set of range-angle-intensity vectors: $\zbf\in \mathbb{R}_+\times[-\pi,\pi)\times\mathbb{R}_+$. While intensity image source data comes from three-dimensional observations, images only contain one angle: either bearing or elevation.\footnote{The associated angle is bearing when the angle sweeps over the $x$-$y$ plane and elevation when sweeping over $x$-$z$.} Therefore, the robot's goal is to reconstruct the underlying 3D coordinates, $\pbf^{(\Rcal)}$, by associating data from multiple images where both $\theta$ and $\phi$ are present.

\begin{figure}[t]
\centering
\includegraphics[width=\columnwidth]{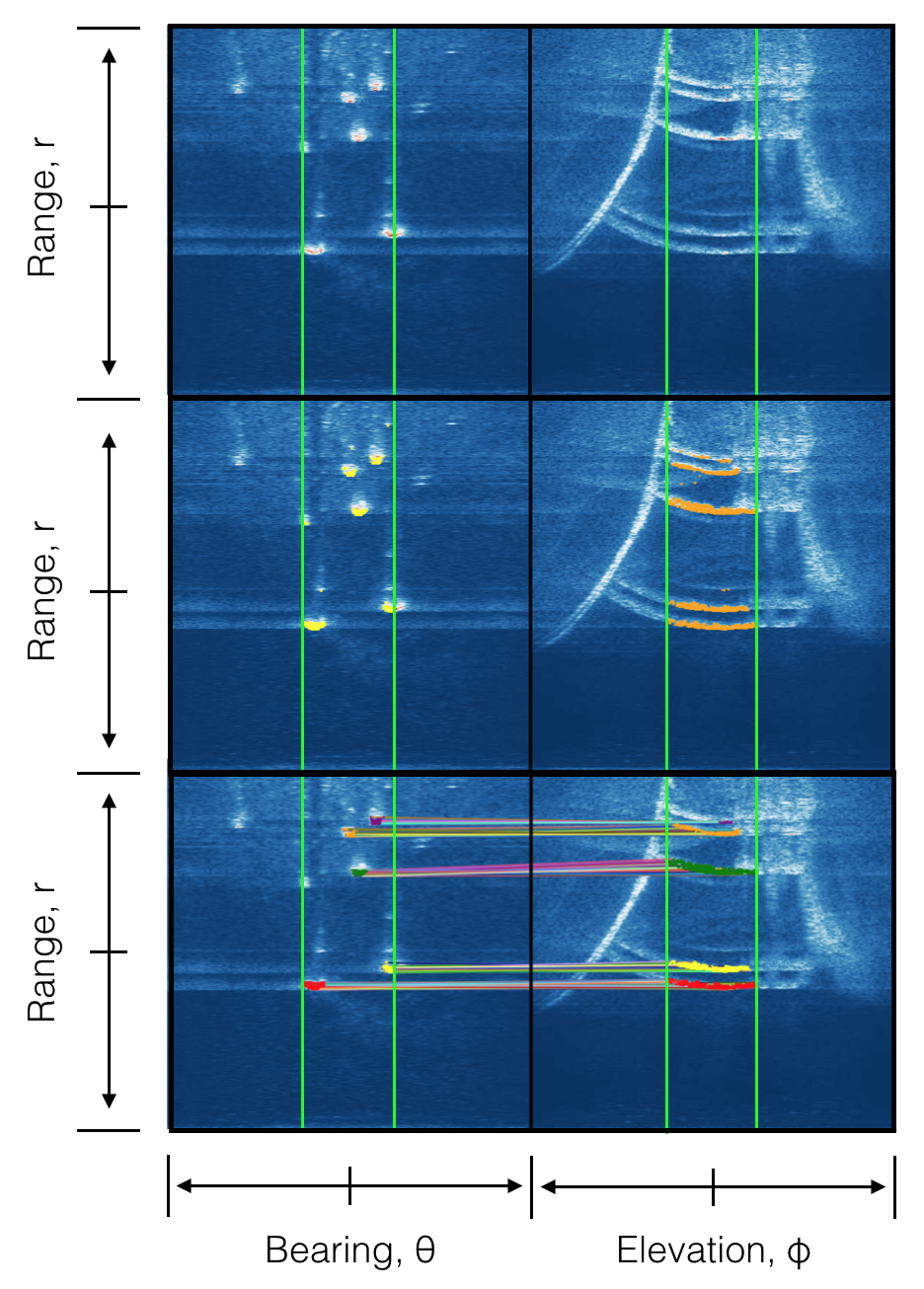}
\caption{\textbf{Architecture overview with sample images.} The left column shows the horizontal sonar with the right showing the vertical sonar. The top row shows raw images, with green lines denoting the overlapping area pictured in Fig. 1(b). Features are shown in the middle row. The bottom row shows matched clusters in color and lines drawn between matched features.}
\vspace{-5mm}
\label{fig:4}
\end{figure}

\subsection{Data Association}
This paper studies methods for data association. We assume that the robot is equipped with two forward-looking acoustic sensors. The sensors are mounted such that their fields of view overlap and permit $\theta$ and $\phi$ to be simultaneously observed. This implies that with a proper calibration, two points from each image correspond to the same object location $\pbf^{(\Rcal)}$. We denote these points as
\begin{align}
    \zbf^{(h)} &= (R^{(h)},\theta,\gamma^{(h)})^\top, &\zbf^{(v)} &= (R^{(v)},\phi,\gamma^{(v)})^\top.
\end{align}
The horizontal sensor, $h$, compresses measurements in the $x$-$y$ plane of $\Rcal$, whereas the vertical sensor, $v$, compresses points in  $x$-$z$. Their associated images sets are denoted
\begin{align*}
    \Zcal^{(h)} &= \{\zbf^{(h)}_1,\cdots, \zbf^{(h)}_{N}\}  ,& 
    \Zcal^{(v)} &= \{\zbf^{(v)}_1,\cdots, \zbf^{(v)}_{N}\},   
\end{align*}
where $N\in\mathbb{N}$ represents the number of observations from each sensor.

Given the intensity images $\Zcal^{(h)}, \Zcal^{(v)}$, we formalize the data association problem as vertex matching in a bipartite graph $\Gcal=(\Vcal,\Ecal)$. The vertices $\Vcal = \Zcal^{(h)} \cup \Zcal^{(v)}$ contain all observed intensity points, and the sets
\begin{align*}
	\Ecal_i = \{(\zbf^{(h)}_i,\zbf^{(v)}_1),\cdots, (\zbf^{(h)}_i,\zbf^{(v)}_N)\},	
\end{align*}
define all realizable associations between points in the horizontal image and points in the vertical image. Their union defines the total edge set $\Ecal = \cup_{i=1}^N\Ecal_i$. Solutions are obtained by finding a set $\Scal\subset\Ecal$ such that
\begin{align}
    \Scal = \bigcup_{\Ecal_i\in\Ecal}\argmin_{(\zbf^{(h)}_i,\zbf^{(v)}_j) \in \Ecal_i}\Lcal(\zbf^{(h)}_i,\zbf^{(v)}_j),
\end{align}
where $\Lcal(\zbf_i,\zbf_j)$ denotes the loss between features. Additionally we require the association to be bijective, in that for any two edges $(\zbf^{(h)}_i,\zbf^{(v)}_i)$, $(\zbf^{(h)}_j,\zbf^{(v)}_j)\in\Scal$ it must be that $\zbf^{(v)}_i\neq\zbf^{(v)}_j$. We estimate $\pbf^{(\Rcal)}$ from Eq \eqref{eq:tx_to_cartesian}, using the fused spherical coordinates
\begin{align}
    \hat{\pbf}^{(\Rcal)}=\left(\frac{R^{(h)}+R^{(v)}}{2},\theta^{(h)},\phi^{(v)} \right)^\top.
    \label{eq:goal}
\end{align} 
Here we use the empirical mean of ranges and the angular values from the resulting association.

\subsection{3D Reconstruction}
Given a set of 3D points $\widehat{\Pcal}^{(\Rcal)} = \{\hat{\pbf}^{(\Rcal)}_i\}_{i=1}^M$, we can complete the reconstruction by mapping these into a fixed inertial frame $\Ical$. This is accomplished with the linear transformation $\Tbf \in \mathbb{R}^{3\times 3}$. When applied to the set of all points, the result is a point cloud which we call \textit{the map}:
\begin{align}
    \widehat{\Pcal}^{(\Ical)} = \{\hat{\pbf}^{(\Ical)}  | \hat{\pbf}^{(\Ical)} = \Tbf \hat{\pbf}^{(\Rcal)} \ \forall \  \hat{\pbf}^{(\Rcal)} \in  \widehat{\Pcal}^{(\Rcal)}\}.
\end{align}


\section{Proposed Algorithm}\label{sec:alg}
Here we describe our proposed methodology for identifying feature correspondences across concurrent orthogonal sonar images (summarized in Figures \ref{fig:fig3} and \ref{fig:4}). The fundamental goal of this pipeline is to associate range measurements from orthogonal, overlapping vantage points that are each lacking a single dimension in their respective spherical coordinate frames. With a set of matched features, the algorithm output (per Eq. (\ref{eq:goal})) is a set of fully defined points in 3D Euclidean space.

\subsection{Feature Extraction}
In our acoustic imagery we do not process every pixel, as not all pixels represent meaningful returns. We must first identify which pixels belong to surfaces in the scene, and to do this, we apply feature extraction to each acoustic image.  

To extract features we use the constant false alarm rate (CFAR) technique \cite{Richards-2005}. This class of algorithm has shown utility in processing radar images \cite {Richards-2005}, \cite{El-Darymli-2018} as well as side-scan sonar images \cite{Acosta-2015}, which are similarly noisy sensing modalities. We have found it to be the most effective feature detector in practice for reliably and consistently eliminating the second returns that regularly appear in sonar imagery. 

\begin{figure}
\centering
\includegraphics[height= 2cm]{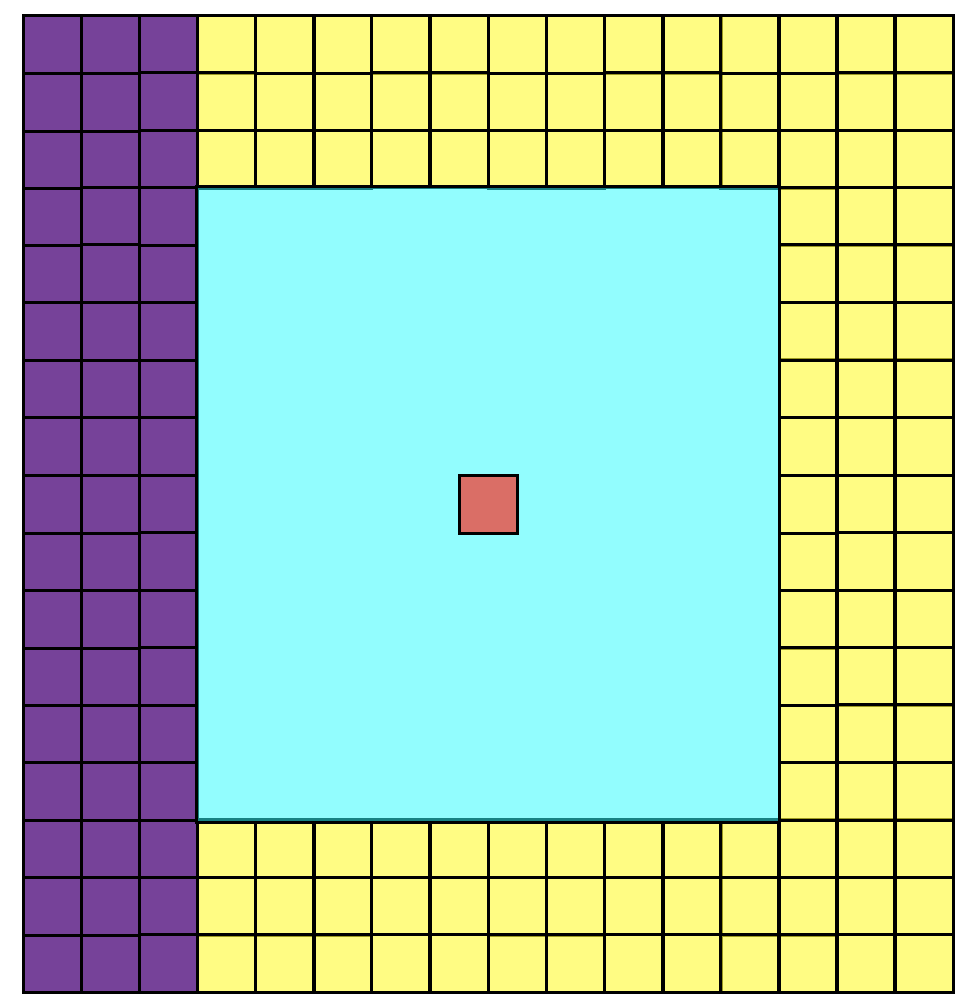}
\includegraphics[height= 2cm]{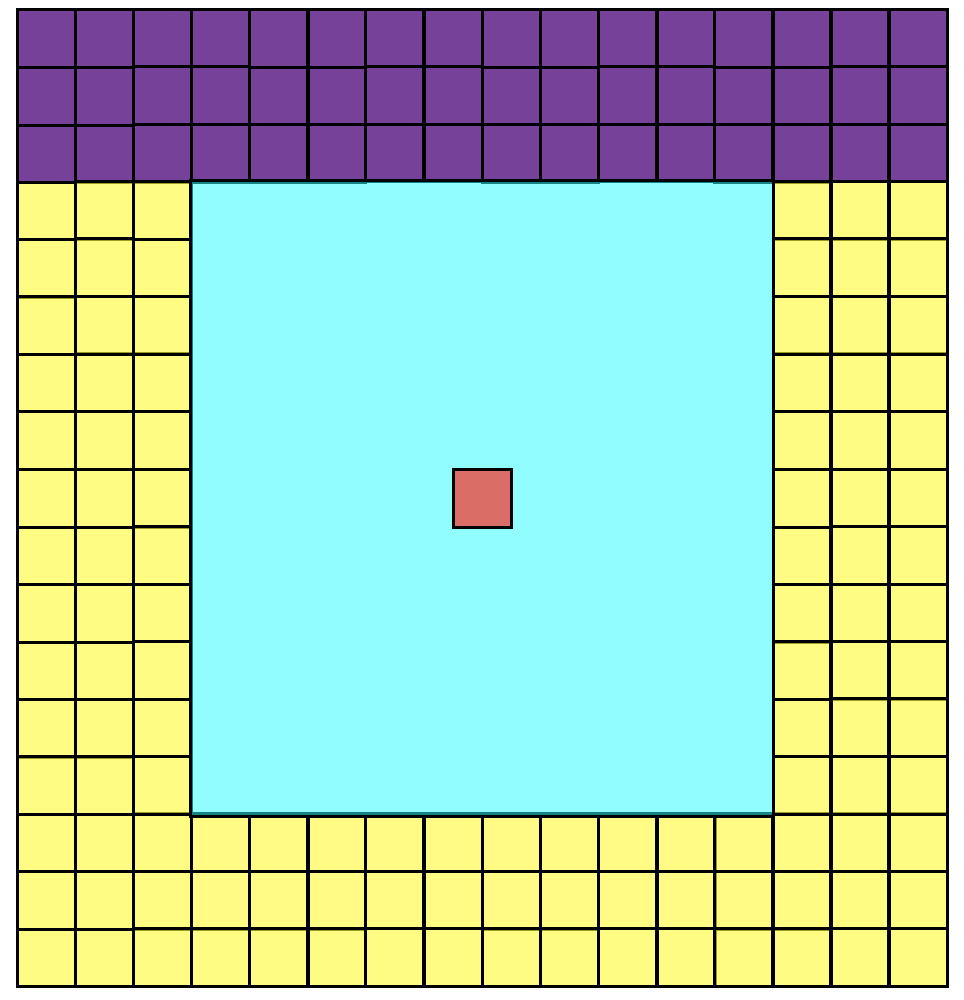}
\includegraphics[height= 2cm]{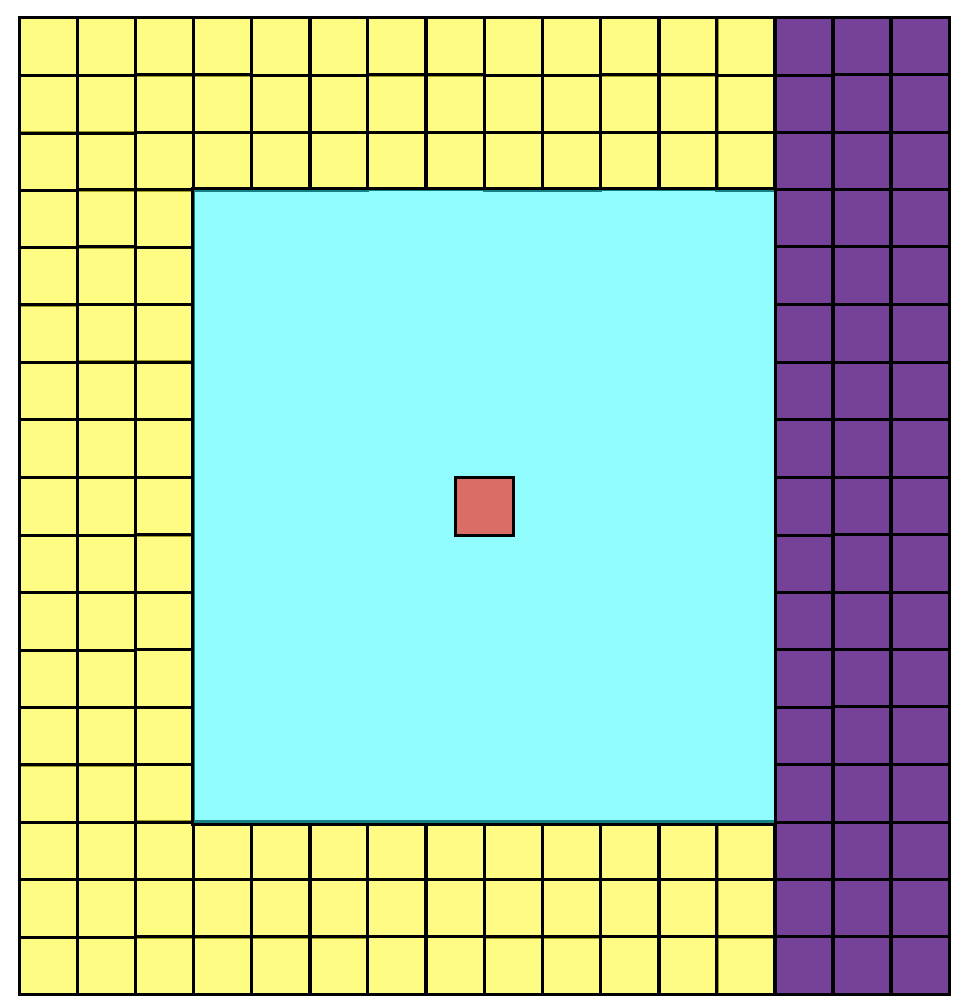}
\includegraphics[height= 2cm]{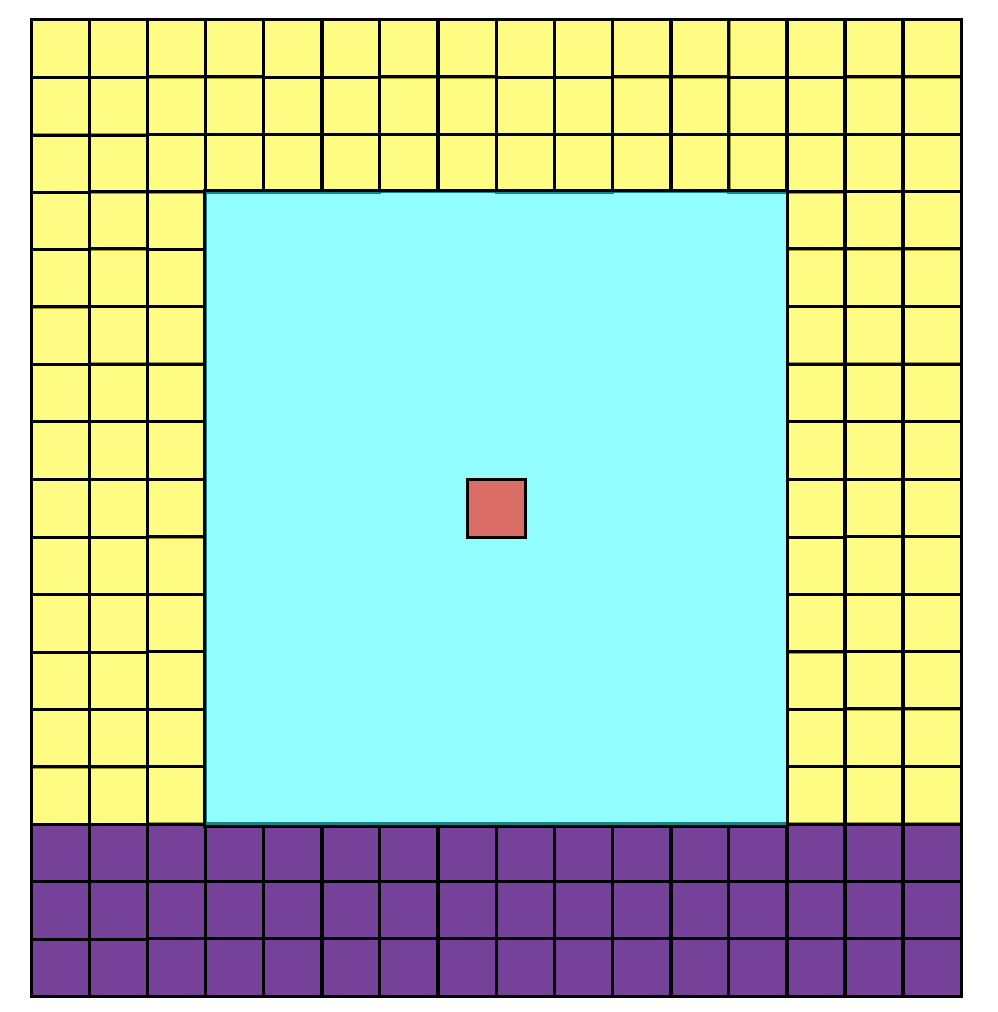}
\caption{\textbf{SOCA-CFAR overview.} Purple cells show the training cells, blue the guard cells and red the cell under test. }
\vspace{-5mm}
\label{fig:5}
\end{figure}

CFAR uses a simple threshold to determine if a pixel in a given image is a contact or not a contact. However, it produces a \textit{dynamic} threshold by computing a noise estimate for the area around the cell under test via cell averaging. The technique is sensitive to multiple targets in the image, especially when the noise estimate includes other positive contacts. It is for this reason we utilize a variant known as ``smallest of cell averages" (SOCA-CFAR) \cite{El-Darymli-2018}. SOCA-CFAR computes four noise estimates and utilizes the smallest of the four. This estimate is expressed in Eq. (\ref{eq:sum}), with $\mathbf{x_m}$ as a training cell, $\mathbf{N}$ as the number of training cells in that quadrant and $\mathbf{\mu}$ as the estimate of noise. This process is shown in Fig. \ref{fig:5}. We note that when averages are computed (purple cells), a layer of guard cells (blue) is wrapped around the cell under test to prevent portions of the signal from leaking into the noise estimate. Next, the detection constant is computed in Eq. (\ref{eq:noise}), with $\mathbf{\alpha}$ as the detection constant. $\mathbf{N}$ is once again the number of cells in the quadrant and $\mathbf{P_{fa}}$ is the specified false alarm rate. The threshold, $\mathbf{\beta}$ can be computed using Eq. (\ref{eq:beta}), with $\mathbf{\mu_{min}}$, the minimum of the computed quadrant averages.  The result of this step in the algorithm is two sets of contacts, $\Scal^{(h)}_t$ from the horizontal sonar and $\Scal^{(v)}_t$ from the vertical. Note that at each time step, there are two sonar images; these images are independently analyzed for features. A sample pair of sonar images with CFAR points is shown in the 2nd row of Fig. \ref{fig:4}. 

\vspace{-5mm}

\begin{align}
\mu= \frac{1}{N} \Sigma_{m=1}^{N}x_m
\label{eq:sum}
\end{align}
\begin{align}
\alpha = N (P_{fa}^{-1/N} - 1 )
\label{eq:noise}
\end{align}
\begin{align}
\beta = \mu_{min} \alpha
\label{eq:beta}
\end{align}

\vspace{-3mm}

\subsection{Clustering and Cluster Association}

Once features are identified in an image pair, these features require an additional constraint for robust association.  To minimize the number of extraneous matches, we can take advantage of the fact that each cluster represents a surface in view.  By first clustering the extracted features and then comparing features between matched clusters only, we can create a significantly more robust pipeline.  

Clustering is performed using the Density-Based Spatial Clustering of Applications with Noise (DBSCAN) \cite{Ester-1996} algorithm on both sets of features $\Scal^{(h)}_t$ and $\Scal^{(v)}_t$. DBSCAN is selected because, in each image, the number of clusters is unknown, and this approach does not require knowledge of the scene a priori. DBSCAN works by iterating through the data, catalogging all core points with more than \textit{minSamples} neighbors lying within a radius $\epsilon$, which, along with their neighbors, form our initial clusters. All other unassigned points lying within $\epsilon$ of a cluster are assigned to that cluster. 
The result of this process is two sets of clusters: $\Ccal^{(h)}_t$ from the horizontal sonar and $\Ccal^{(v)}_t$ from the vertical sonar. 

Next, to match clusters across orthogonal sonars, we compute four descriptors and then minimize a cost function to associate clusters. Each cluster is defined by its mean range $\mathbf{\mu}$, variance in range $\mathbf{\sigma^2}$, and min and max in range, shown in Eq. (\ref{eq:cluster_des}).  Each cluster $\cbf^{(h)}_t$ is assigned to the cluster in $\Ccal^{(v)}_t$ that minimizes cost function (\ref{eq:cluster_cost}):

\begin{equation}
\cbf_t = \begin{bmatrix} \mu \ \sigma^2 \ r_{min} \ r_{max} \end{bmatrix}^\top,
\label{eq:cluster_des}
\end{equation}
\begin{equation}
    \Lcal(\cbf_t^{(h)},\cbf_t^{(h)}) = || \cbf_t^{(h)} -\cbf_t^{(v)}||_2.  
    \label{eq:cluster_cost}
\end{equation}

\subsection{Feature Association}

Following the clustering algorithm output, the next stage is to match individual features within our matched clusters. A feature can only be matched to another feature if they belong to the same cluster, which greatly reduces the potential for extraneous feature matches.

To match features, we once again adopt the descriptor and cost function paradigm. Each feature is defined by range, $\mathbf r$ intensity $\mathbf \gamma$, and a mean intensity kernel.  We consider two mean terms for each feature, which are broken into two axes.  $\mathbf{\mu_x}$ is the mean intensity of $\mathbf i$ points right and $\mathbf i$ points left of the feature.  $\mathbf{\mu_y}$ is the mean intensity of $\mathbf i$ points above and $\mathbf i$ points below the feature, in image coordinates.  

The reasoning for this is straightforward; the body of work addressing elevation angle estimation in sonar imagery relies on the relationship between incident angle and intensity.  Since our goal is to match similar measurements, we leverage this relationship by hypothesizing that not only should similar measurements have similar range, but also intensities, because of their similar incident angles. However, due to the noise in acoustic images, we adopt not only range and intensity as feature descriptors, but also local averages of intensity. Note that feature descriptors in Eqs. (\ref{eq:des_1}) and (\ref{eq:des_2}) have mismatched axes in their $\mathbf{\mu}$ terms, which is due to the orthogonality of the images: 

\begin{equation}
\zbf_i^{(h)} = \begin{bmatrix} r \ \gamma \ \mu_x \ \mu_y \ \end{bmatrix}^\top 
\label{eq:des_1}
\end{equation}
\begin{equation}
\zbf_j^{(v)} = \begin{bmatrix} r \ \gamma \ \mu_y \ \mu_x \ \end{bmatrix}^\top 
\label{eq:des_2}
\end{equation}
\begin{equation}
    \Lcal(\zbf_i^{(h)},\zbf_j^{(v)}) = || \zbf_{i}^{(h)} -\zbf_{j}^{(v)}||_2.
    \label{eq:feature_cost}
\end{equation}

We next compute the minimum-loss association among the features residing in previously associated clusters. When carrying out this process, we only allow any feature to be matched with a single other feature. Moreover, before evaluating cost we \textit{normalize} all components by subtracting from each feature the min image intensity and dividing by the difference between min and max. These min and max values are computed for each image, at every timestep. 

In our implementation, we have developed two versions of this process; the first is a ``brute force" method in which all possible correspondences for a given cluster are tested for each feature, and the lowest cost is used provided it is below a designated threshold. Alternatively, we take a fixed number of random samples from a cluster and again use the lowest cost, provided it is below our threshold. These two versions of our proposed algorithm will later be quantitatively compared. 

\begin{figure}[t]
  \centering
  \subfloat[Isometric view]{\includegraphics[width=0.4\columnwidth]{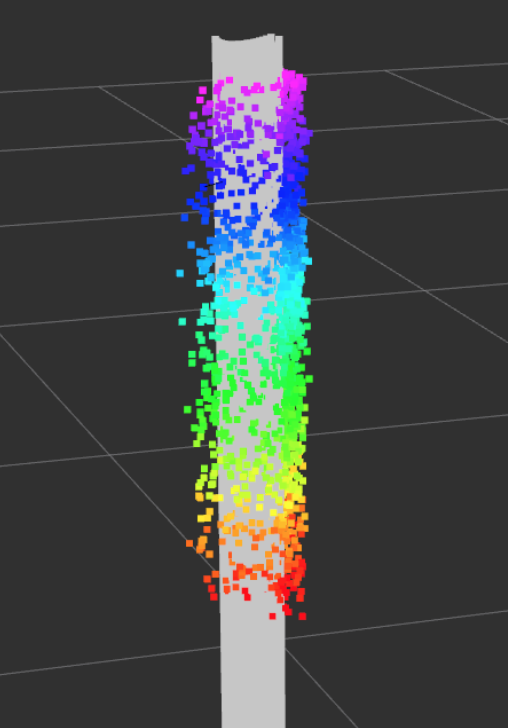}\label{fig:f1}}
  \hfill
  \subfloat[Top down view]{\includegraphics[width=0.4\columnwidth]{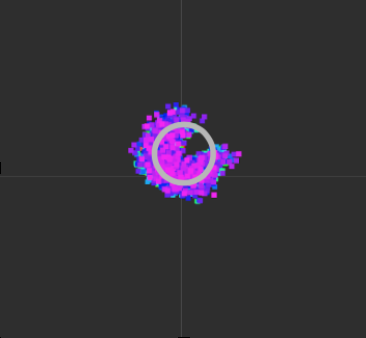}\label{fig:f2}}
  \caption{\textbf{Results from pier piling mockup.} A model of the structure's true dimensions is shown in gray (outer diameter is 9cm), with multi-colored points depicting algorithm output. Sonars were operated at 5m range, and results from our algorithm's ``fast, clustering'' configuration are shown. Colors indicate height. 
  }
  \vspace{-4mm}
  \label{fig:figure6}
\end{figure}

\section{Experiments and Results}

In this section, we will examine three experiments and provide an analysis of their outcomes.  We will discuss the specifics of our experimental setup and the hardware used to validate the proposed perceptual framework of Fig. \ref{fig:fig3}.  

\subsection{Hardware Overview}
In order to collect data for evaluation, our heavy-configuration BlueROV2  underwater robot was deployed with a customized sensor payload that includes a camera, a Rowe SeaPilot Doppler velocity log (DVL), a VectorNav VN100 inertial measurement unit (IMU), and a Bar30 pressure sensor. In addition, two forward-looking multi-beam imaging sonars were used (selected based on our available resources for these experiments); Oculus M750d and Oculus M1200d, which were operated in their low-frequency, wide-aperture modes at 750kHz and 1.2Mhz, respectively. In this mode, both sensors have a vertical aperture of 20$^{\circ}$ and a horizontal aperture of 130$^{\circ}$. 
The M750d was mounted in the horizontal configuration, shown in red in Fig. \ref{fig:1}(b), and the M1200d was mounted in the vertical configuration, shown in blue in Fig. \ref{fig:1}(b). These sonars return slightly different intensity magnitudes due to their frequencies, however at each time step all feature intensity values extracted from an image are normalized, as described in Section IV.C.

The Robot Operating System (ROS) \cite{Quigley-2009} was used to operate the vehicle, and to log data from a topside computer.

\subsection{Overlapping Area Considerations}
When extracting features, we take a highly conservative approach to ensure that features are only extracted from the region of overlap between the sonars. We only extract features from the area inside the vertical aperture of each sonar's orthogonal companion, shown in green in Fig. \ref{fig:4}. This conservative approach ensures no features are extracted outside of the overlapping area depicted in Fig. \ref{fig:1}(b). 

\subsection{Compensating for Sonar Misalignment}
Before any data association can take place, we must transform our vertical sonar features $\Scal^{(v)}_t$ to the horizontal sonar frame. In our experimental setup, there is a  10cm vertical offset between the sonar coordinate frames, shown in Fig. \ref{fig:1}(a). This is accounted for by applying Eq. (\ref{eq:tx_to_cartesian}) to transform our features in $\Scal^{(v)}_t$ to Cartesian coordinates, translate them a few centimeters downward to the horizontal sonar coordinate frame, and then apply the inverse of Eq. (\ref{eq:tx_to_cartesian}) to transform the features back into spherical coordinates for association. To apply Eq. (\ref{eq:tx_to_cartesian}), an elevation angle of zero is assumed for all points in the vertical sonar image (this assumption is used to facilitate data association only). 

\subsection{Error Metrics}
To compare different parameterizations of our method against each other and a benchmark, we utilize two error metrics. Firstly we compute mean absolute error (MAE); this is calculated by comparing the generated point cloud to a CAD model of the object. Secondly, we compute the root mean squared error (RMSE); once again, this is generated by comparing the point cloud to the CAD model.
For our tank experiments, to capture the transformation between the ROV frame and the CAD model with high accuracy, the objects in our sonar imagery are hand segmented to identify both the object coordinate frame origin (located at the center of the pipe comprising all structures), and its rotation in yaw relative to the ROV. The remaining angles in the transformation are obtained from the ROV's IMU.

\subsection{Simple Object Reconstruction}
In this first experiment, a cylindrical piling mock-up is submerged in our test tank, and a sequence of data is collected with our sonar pair oriented at a grazing angle of 20$^{\circ}$ below horizontal. This grazing angle is used in order to facilitate benchmarking, as the current state of the art \cite{Westman-2019} assumes this problem structure. When collecting data, the ROV is piloted in a circular-segment orbit pattern around the structure. During this flight pattern, the structure is held at a similar range and bearing while the ROV traverses to port or starboard. Data collection occurs at a fixed depth.  

The purpose of this experiment is to evaluate the performance of the proposed algorithm on a simple structure.  Additionally, this experiment allows us to evaluate our algorithm against the current state of the art, as \cite{Westman-2019} shows several experiments reconstructing similar objects. During the comparison, it is essential to note that we use SOCA-CFAR feature extraction in conjunction with our implementation of \cite{Westman-2019} to ensure that all systems run on the same inputs. 

While the goal of our algorithm is to move toward the reconstruction of arbitrary objects with wide-aperture multi-beam sonar, we are acutely aware that the algorithm we are proposing requires a second sonar. This experiment serves to show that performance does not degrade in the case of  objects that can be successfully reconstructed with a single sonar. We compare four variations of our algorithm against \cite{Westman-2019}, as it achieves the best performance in our tank among the suitable algorithms in the literature. Additionally, we posit that \cite{Westman-2019} fairly represents the foundational works of \cite{Aykin-2013}, \cite{Aykin-2013-1}, building upon them and offering broader applicability. 

We compare four configurations of our algorithm to show performance gains and trade-offs for different versions quantitatively. Firstly we evaluate the introduction of clusters as feature correspondence constraints. Secondly, we compare the ``brute force" approach in which all feature combinations are checked via (\ref{eq:feature_cost}), and a version in which ten random samples from a feature's corresponding cluster are checked, with the best adopted if below a designated threshold. This second version runs in real-time over all data gathered, while the brute force method runs significantly slower. We also note that \cite{Westman-2019} runs in real-time in these experiments. The feature correspondence threshold is the same for all methods and experiments provided in this paper; it is set to 0.1.  

Experimental results are shown in Table \ref{eq:ep1} and Fig. \ref{fig:figure6}; our proposed algorithm performs comparably to the current state of the art, and moreover, the results from our implementation of \cite{Westman-2019} are in line with those shown in the original paper. Any variation in performance relative to \cite{Westman-2019} can be attributed to the fact that the sonar used in our implementation has half the angular resolution as in \cite{Westman-2019}, and we analyze raw point clouds rather than filtered surface meshes.  

When analyzing the results of different configurations of our proposed algorithm, it is not surprising that in this simple case, with only a single cylindrical piling in view, the cluster constraints provide little added value. Additionally, the trade-off between fast and brute force methods is evident, with a slight loss of accuracy in exchange for real-time viability.

\begin{table}[t]
\centering
\begin{tabular}{ccc}
\toprule
Algorithm & MAE (cm) & RMSE (cm) \\
\midrule
Westman and Kaess \cite{Westman-2019} & 2.20 & 2.64  \\
Brute Force Without Clustering & 2.16 & 2.53  \\
Fast Without Clustering & 2.34 & 2.76  \\
Brute Force With Clustering & 2.27 & 2.70  \\
Fast With Clustering & 2.35 & 2.83  \\

\toprule
\end{tabular}
\caption{A summary of reconstruction performance corresponding to the reconstruction of a single piling (pictured in Fig. 6).}
\vspace{-4mm}
\label{eq:ep1}
\end{table}

\subsection{Complex Object Reconstruction}

Recall that to reconstruct objects using a single wide-aperture imaging sonar, the framework proposed in \cite{Westman-2019} must make two critical assumptions; the first being that the range to an object increases or decreases monotonically with elevation angle. The consequence of an object violating this assumption is the inability to reconstruct the geometry accurately. Moreover, \cite{Westman-2019} states that ``a violation of this assumption would cause a self-occlusion, and the corresponding pixels would presumably not be classified as surface pixels by the frontend of our algorithm." The second key assumption is that the sonar is oriented at a downward grazing angle, which enables identification of an object's leading edge.  
In our proposed framework, we require no assumptions about the geometry of the objects in view, nor do we require the sensor at a grazing angle.

In this experiment, we test our proposed algorithm on a mock-up of a critical piece of subsea infrastructure, the blow out preventer (BOP) - approximated by a rectangular object mounted on a cylinder. BOPs sit on the seafloor at the wellhead during offshore drilling and are increasingly subject to regulatory scrutiny and industry monitoring requirements. 
Critically though, like many other subsea assets, the vertical cross-section of this object does not conform to the aforementioned geometric assumptions. 

Once again, data is collected by piloting the ROV in a circular-segment orbit around a portion of the structure (low clearances in our tank prevent full circumnavigation of the structure).  Recall that this flight pattern keeps the structure at a similar range and bearing while the ROV traverses to port or starboard. In this experiment, however, the sonar is not at a grazing angle, configured instead for maximum situational awareness. Unlike in the other experiments in this paper, data is collected at a variety of depths.

\begin{figure}[t]
\centering
\subfloat[Water Tank Setup \label{fig:6a}]{\includegraphics[height=3.3cm]{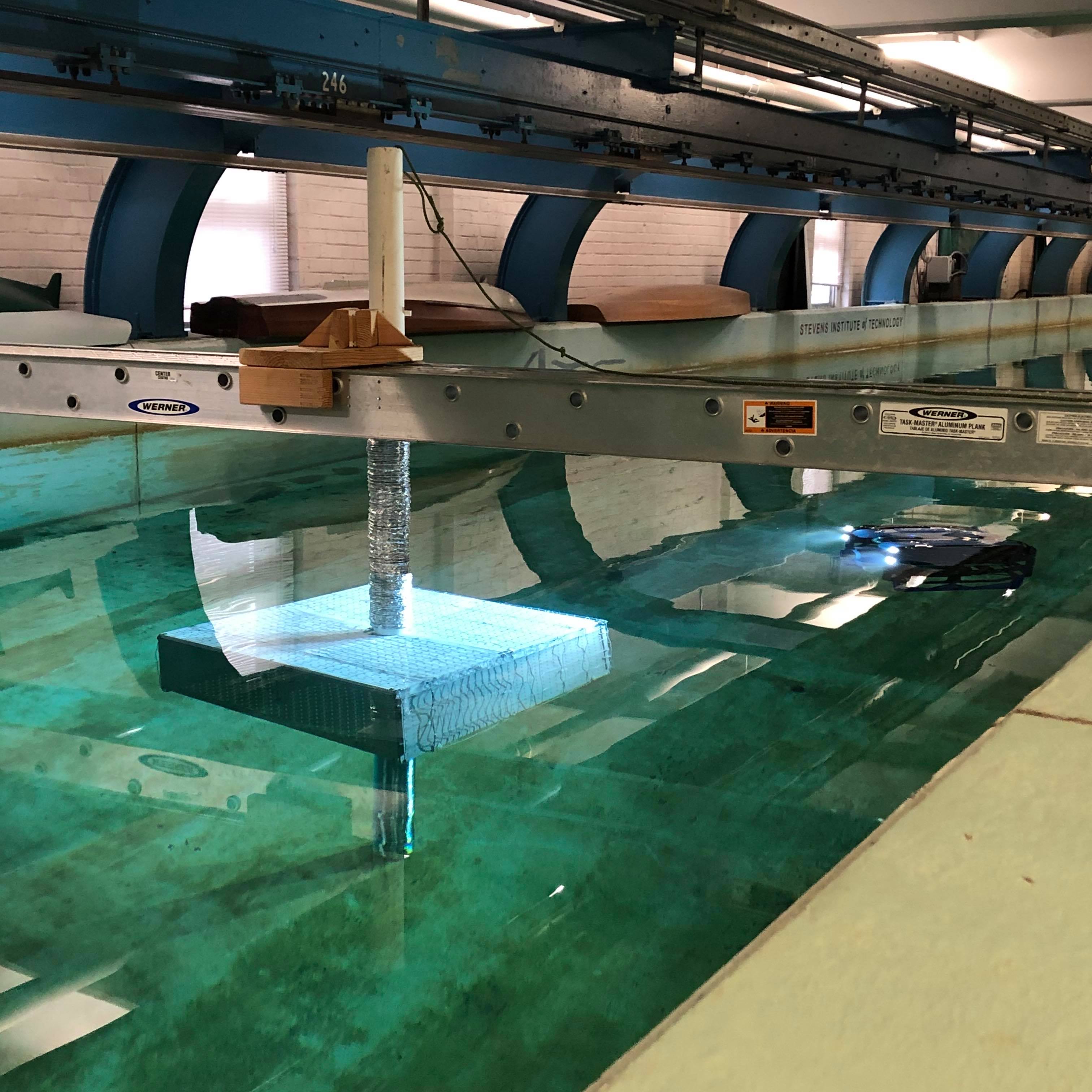}}\
\subfloat[Isometric view \label{fig:6a}]{\includegraphics[height=3.3cm]{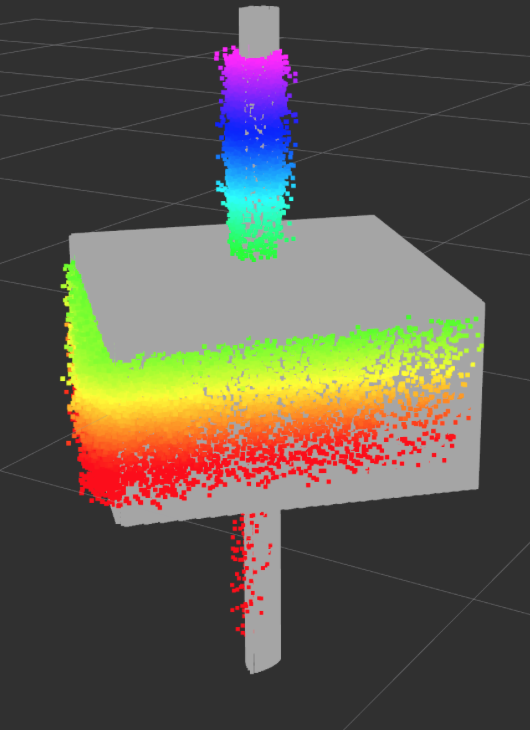}}\

\subfloat[Top View \label{fig:6b}]{\includegraphics[height=3.3cm]{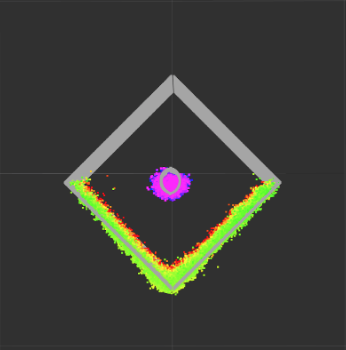}}\
\subfloat[Side View\label{fig:6c}]{\includegraphics[height=3.3cm]{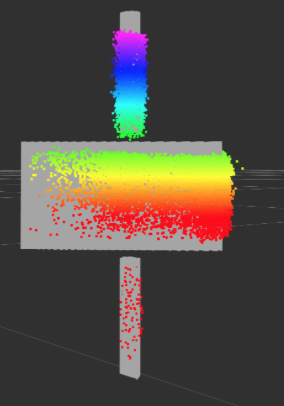}}\

\caption{\textbf{Results from blowout preventer mockup.} Gray shows the true structure geometry (a 9cm OD pipe, with an 82 x 82 x 45cm box, whose center is 1.2m from the bottom of pipe), and in (c), the top of the structure is not shown for ease of visualization. Sonars were operated at 5m range, and results from our algorithm's ``fast, clustering'' configuration are shown. Color indicates height.
}
\label{fig:tank_cmp}
\vspace{-5mm}
\end{figure}

A summary of the results is provided in Table \ref{eq:table2} and Fig. \ref{fig:tank_cmp}. 
Not only is our algorithm able to provide a realistic reconstruction of the object, but it can do so within 6cm of its true geometry. In analyzing the results of this experiment, the benefits of clustering as a constraint in feature association are evident; clustering dramatically improves both MAE and RMSE. Again the trade-off between brute force and fast feature-matching is evident; a small decrease in performance is required in order to run in real-time. This trend is not evident without clustering, and the reason for this is not known for certain, but the random guessing associated with the fast version may be acting as a filter. 
The reconstruction's appearance in Fig. \ref{fig:tank_cmp} is realistic, with no evident outliers, particularly in the vertical axis. While it is unfortunate our algorithm is unable to reconstruct the top surface of the rectangular object, this was not surprising given the object's sharp angles, and the occlusions they present. From the vantage points examined, the top rectangular surface of the structure was imaged at much lower intensity than its front edges. 
Video playback of our algorithm's mapping of the BOP structure is provided in our video attachment.
We have omitted a direct comparison with \cite{Westman-2019} for this structure in an effort not to misrepresent their work, in a modality they explicitly state their algorithm is not intended for.

\begin{figure*}[!ht]
  \centering
  \subfloat[Outdoor Pier Pilings]{\includegraphics[width=0.2455\textwidth]{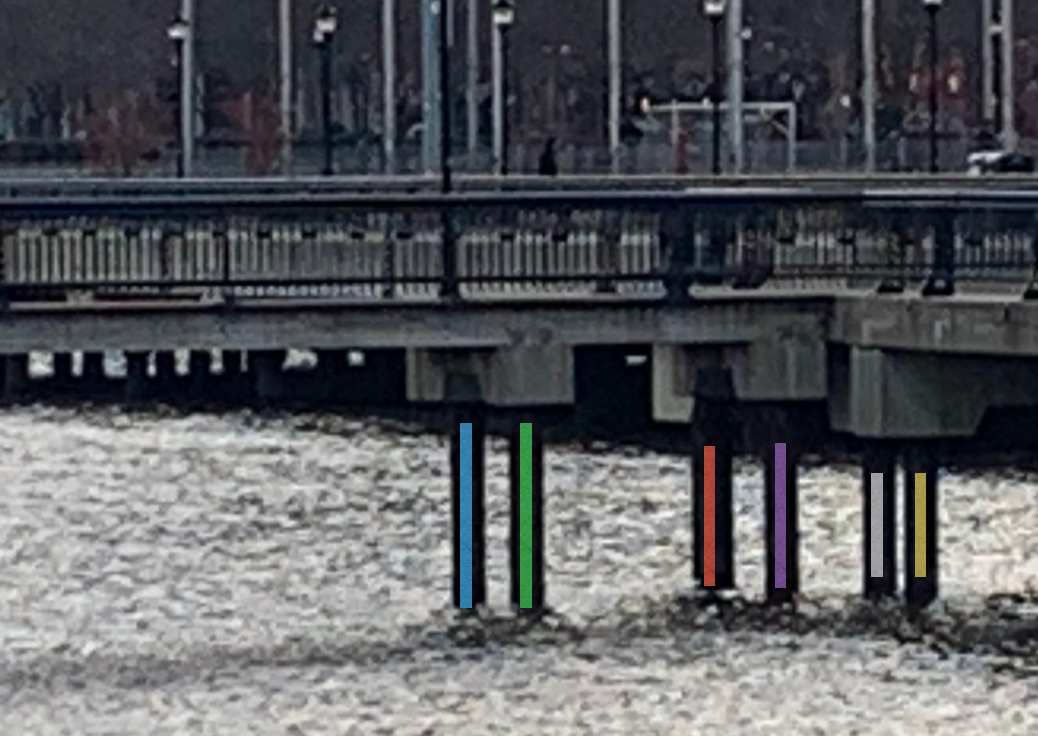}\label{fig:f1}}
  \subfloat[Piling Reconstruction]{\includegraphics[width=0.26\textwidth]{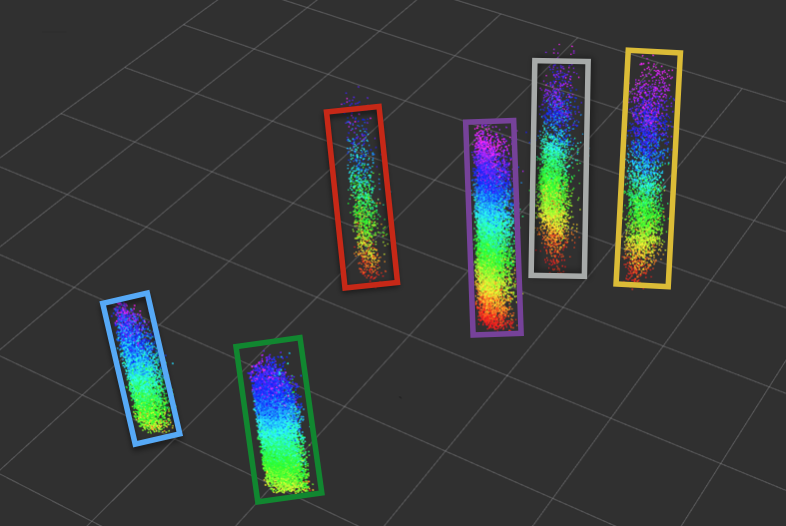}\label{fig:f2}}
  \caption{Hudson River pier pilings at left with reconstruction at right (shown also in Fig. 1(c)). Data was collected at a fixed depth, with sonars operated at 10m range, and results from our algorithm's ``fast, clustering'' configuration are shown. Point colors indicate height.}
  \vspace{-5mm}
  \label{fig:outdoor}
\end{figure*}

\begin{table}[h!]
\vspace{-1mm}
\centering
\begin{tabular}{ccc}
\toprule
Algorithm & MAE (cm) & RMSE (cm) \\
\midrule
Brute Force Without Clustering & 62.29 & 69.39  \\
Fast Without Clustering & 30.24 & 45.91  \\
Brute Force With Clustering & 5.31 & 10.06  \\
Fast With Clustering & 5.74 & 12.75  \\

\toprule
\end{tabular}
\caption{A summary of model reconstruction performance corresponding to the reconstruction of the BOP mockup.}
\vspace{-4mm}
\label{eq:table2}
\end{table}

\subsection{Field Based Object Reconstruction}

The final experiment to be presented is a demonstration of the proposed algorithm operating in the field. For this work, our ROV was deployed in the Hudson River at Hoboken, NJ. A short inspection flight was flown at the junction of two piers with six pilings supporting the structure close to each other.  This exercise takes place at a fixed depth and the sonars without any grazing angle. The version of the algorithm used for this demonstration is fast, with clustering.  

The results, shown in Figs. \ref{fig:1}(c) and \ref{fig:outdoor} (and whose sonar images also appear in illustrative Figs. \ref{fig:fig3} and \ref{fig:4}), demonstrate that a robot employing the proposed framework can achieve significant situational awareness, even while operating at a fixed depth. We can recover dense 3D point clouds at every timestep, which are registered here using only iterative closest point (ICP). For the same inspection coverage to be achieved with a profiling sonar, changes in depth would be required. Furthermore, reconstructing these pilings with a single wide-aperture sonar \cite{Westman-2019} would require a grazing angle that would limit situational awareness.
Moreover, if the vehicle is perturbed, an event of significant likelihood given the wakes and currents encountered in this tidal river basin, the requirements of prior algorithms may be violated.  There is another crucial trade-off here: to extract 3D information from the sensors, a reduction in the horizontal field of view to the overlapping area between sonars is required. We believe this is a reasonable trade, though, as Figs. \ref{fig:fig3} and \ref{fig:4} demonstrate that reasonable coverage can be achieved, even with the reduced-size portions of sonar images that overlap, at the sensing range of 10m explored in this experiment.

\section{Conclusions}

In this paper, we have presented a new framework for achieving dense 3D reconstruction of arbitrary scenes using a pair of orthogonally oriented, wide-aperture multi-beam imaging sonars. We have provided a detailed description of a pipeline for robust feature extraction, clustering, and descriptors that facilitate accurate matching across concurrent sonar frames. Further, we have shown quantitatively that our algorithm performs competitively with the state of the art in reconstruction using a single imaging sonar, in simple cases where a comparison is possible. Most importantly, this methodology introduces a new dense sonar mapping capability for complex scenes, and unlike previous work it requires few if any assumptions about the environment, advancing progress toward reconstructing arbitrary geometries and cluttered scenes with wide-aperture multi-beam sonar.  

Future work will focus on using this methodology as a basis for mapping beyond the area of explicit overlap between sonars, and incorporating this framework into a process for robust, underwater 3D active SLAM. 

\vspace{-1mm}

\section*{Acknowledgements}
This research was supported by a grant from Schlumberger Technology Corporation. We thank Timothy Osedach, Stephane Vannuffelen and Arnaud Croux for constructive comments that have improved the quality of this manuscript.
 
{}
\end{document}